\documentclass[pdflatex,sn-mathphys-num]{sn-jnl}

\usepackage{graphicx}%
\usepackage{multirow}%
\usepackage{amsmath,amssymb,amsfonts}%
\usepackage{amsthm}%
\usepackage{mathrsfs}%
\usepackage[title]{appendix}%
\usepackage{textcomp}%
\usepackage{manyfoot}%
\usepackage{booktabs}%
\usepackage{algorithm}%
\usepackage{algorithmicx}%
\usepackage{algpseudocode}%
\usepackage{listings}%
\usepackage[table,xcdraw]{xcolor}
\usepackage{hyperref}

\begin{document}

\title[Article Title]{3D/2D Registration of Angiograms using
Silhouette-based Differentiable Rendering}

\author[1]{\fnm{Taewoong} \sur{Lee}}

\author[1,2]{\fnm{Sarah} \sur{Frisken}}

\author[1,2]{\fnm{Nazim} \sur{Haouchine}}

\affil[1]{\orgname{Harvard University}, \orgaddress{\city{Cambridge}, \state{MA}, \country{USA}}}

\affil[2]{\orgname{Brigham and Women's Hospital}, \orgaddress{\city{Boston}, \state{MA}, \country{USA}}}

\abstract{We present a method for 3D/2D registration of Digital Subtraction Angiography (DSA) images to provide valuable insight into brain hemodynamics and angioarchitecture. Our approach formulates the registration as a pose estimation problem, leveraging both anteroposterior and lateral DSA views and employing differentiable rendering. Preliminary experiments on real and synthetic datasets demonstrate the effectiveness of our method, with both qualitative and quantitative evaluations highlighting its potential for clinical applications. The code is available at \url{https://github.com/taewoonglee17/TwoViewsDSAReg}.}


\maketitle

\section{Introduction}\label{sec1}
3D/2D registration of Digital Subtraction Angiography (DSA) images can bring useful clinical insight into brain hemodynamics and angioarchitecture, as demonstrated in studies such as \cite{frisken2022using,haouchine2021estimation,zhang2023enhancing}. Conventional methods can be categorized into four categories: intensity-based, feature-based, gradient-based, and hybrid \cite{mitrovic20183d}.
Intensity-based methods directly compare voxel/pixel-wise the 3D and 2D images \cite{hipwell2003intensity}, while feature-based methods instead compare features extracted from the 2D image and its corresponding feature in the 3D image like vessel centerline points \cite{feldmar19973d}.
Gradient-based methods use the principle that the 2D gradient reflects the perpendicular component of the 3D gradient relative to the incident ray direction \cite{markelj2008robust}, and hybrid methods combine distinct registration bases for 3D and 2D, such as aligning 2D projections of 3D vessel orientations with 2D intensity gradients using a local neighborhood-based similarity measure \cite{pernus20133d}.
Recent approaches using deep learning have shown promising results, improving both the accuracy and robustness compared to conventional methods. Some of these deep learning methods include using a Spectral Regression model \cite{tang2016similarity}, a neural network using the U-net architecture \cite{hellebrekers2024automated}, a Projective Spatial Transformer (ProST) module \cite{gao2023fully}, and DiffPose \cite{gopalakrishnan2024intraoperative}.
Differentiable rendering has also been extensively used given its practical characteristics when used in conjunction with deep learning, as seen in the differentiable ProST method \cite{gao2023fully}, the differentiable rendering of synthetic X-rays \cite{gopalakrishnan2024intraoperative}, or when using Neural Radiance Fields \cite{fehrentz2024intraoperative}.

\begin{figure}[t]
    \centering
    \href{https://github.com/taewoonglee17/TwoViewsDSAReg/}{%
        \includegraphics[width=0.46\textwidth]{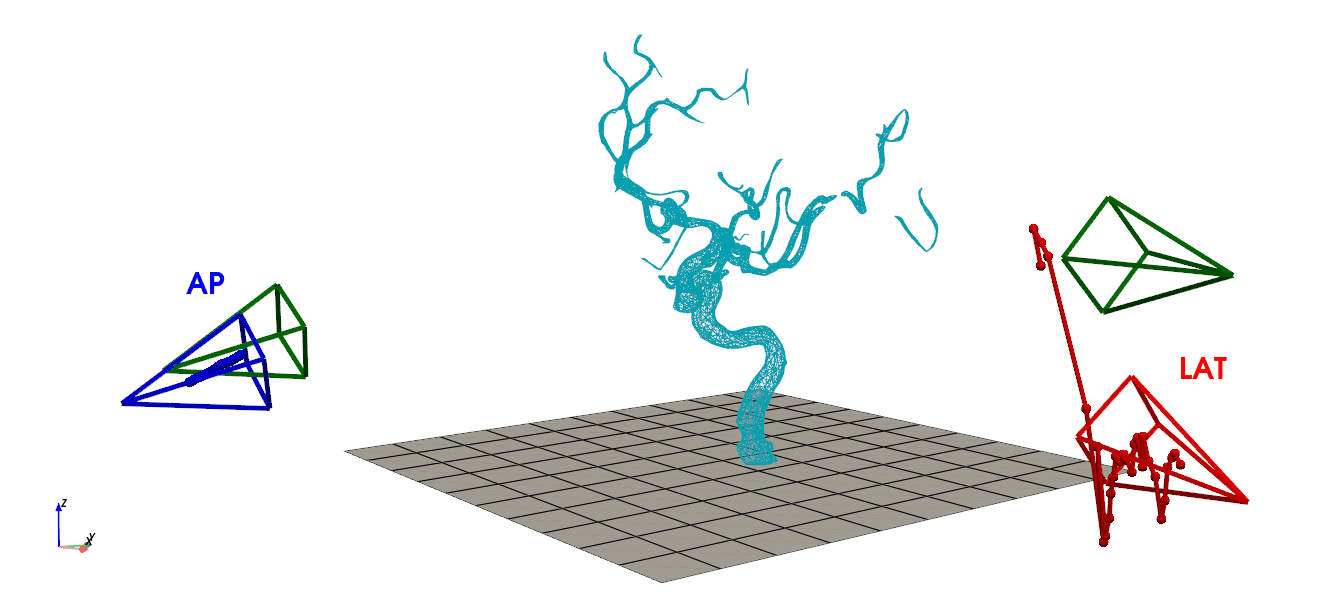}%
        \includegraphics[width=0.26\textwidth]{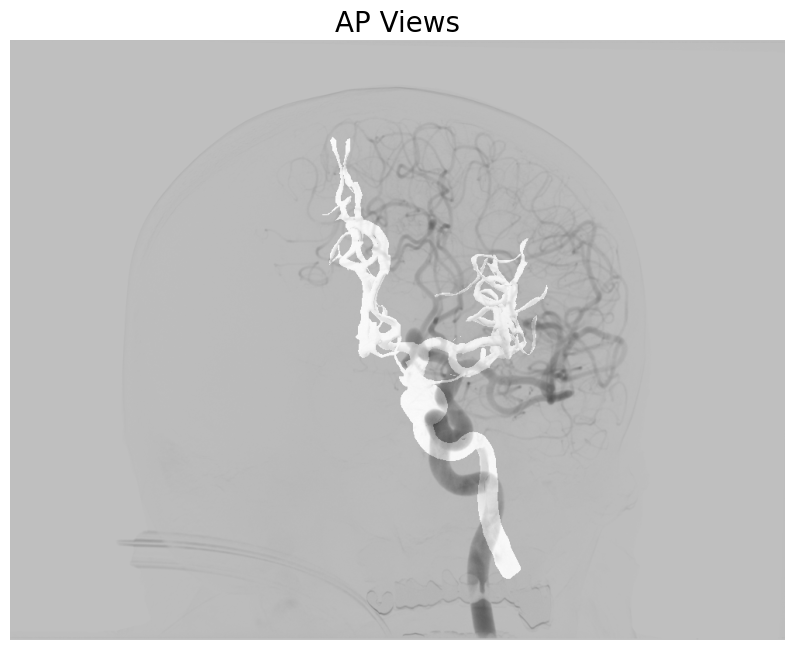}%
        \includegraphics[width=0.26\textwidth]{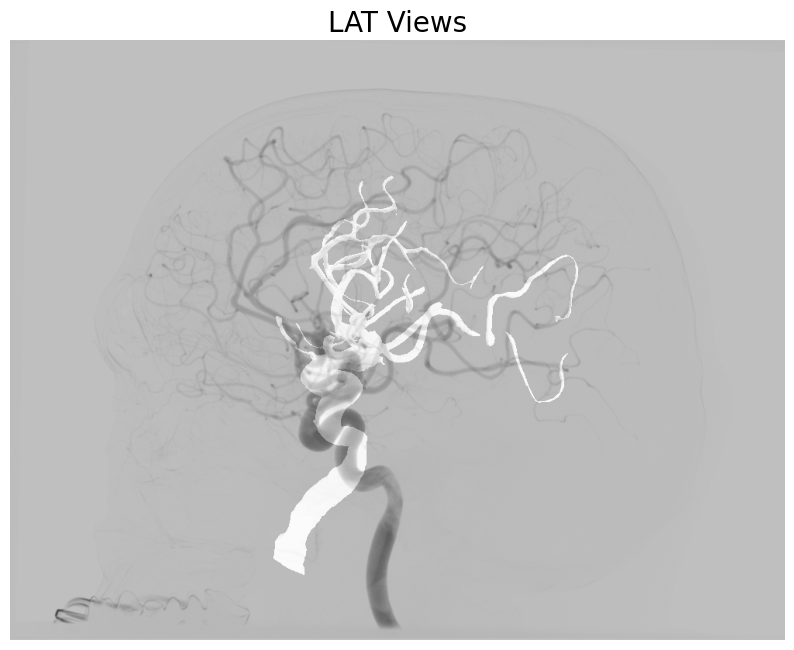}%
    }
\caption{Overview of the proposed silhouette-based differentiable rendering method for 3D/2D angiographic registration using two views. The framework optimizes the alignment of 2D DSA images with the 3D vascular model. \textbf{Click the image to play the video in a browser.}}
\label{fig:overview}
\end{figure}

We propose a novel approach to perform 3D/2D registration of angiograms. We tackle the problem as a pose estimation problem using both anteroposterior (AP) and lateral (LAT) DSA images and optimize with differentiable rendering. We present preliminary results on real and synthesized data with qualitative assessment and quantitative measurements. Our method is illustrated in Figure \ref{fig:overview}.

\section{Method}

\subsection*{Problem Formulation}
Let $\mathbf{I}_\text{AP} \in \mathbb{R}^2$ and $\mathbf{I}_\text{L} \in \mathbb{R}^2$ denote the anteroposterior and lateral 2D DSA images, respectively, acquired by the X-ray beam projectors $\mathbf{P}_\text{AP} \in SE(3)$ and $\mathbf{P}_\text{L} \in SE(3)$. Here, $SE(3)$ denotes the special Euclidean group. Additionally, let $\mathbf{M} \in \mathbb{R}^3$ represent the 3D geometric vasculature, which can be derived from a CTA, MRA, or 3D DSA image.

We seek to determine the poses $\mathbf{P}_\text{AP}$ and $\mathbf{P}_\text{L}$, which minimize a loss function $\mathcal{L}( \{\mathbf{P}_\text{AP}; \mathbf{P}_\text{L}\} | \{\mathbf{I}_\text{AP}; \mathbf{I}_{L}\},  \mathbf{M})$ that quantifies the discrepancy between the pair of 2D DSA images and the 3D mesh $\mathbf{M}$ when the X-ray beam is positioned and oriented according to the each respective pose $\mathbf{P}$.

We approach the problem as an optimization problem in 2D image space to minimize the loss between segmentations of the image pair $\{\mathbf{I}_\text{AP}; \mathbf{I}_{L}\}$ and silhouette images rendered using a differentiable renderer, yielding:
\begin{equation}
  \widehat{\mathbf{P}_\text{AP}}, \widehat{\mathbf{P}_\text{L}}  = \underset{\mathbf{P}_\text{AP},\mathbf{P}_\text{L}}{\mathrm{argmin}} \; \mathcal{L}(\{\mathbf{I}_\text{AP}; \mathbf{I}_{L}\}, f(\{\mathbf{P}_\text{AP}; \mathbf{P}_\text{L}\} | \mathbf{M}))
\label{eq:image_problem}
\end{equation}
where $f$ is the silhouette rendered, differentiable with respect to the poses, allowing for a gradient-based iterative optimization. 

\subsection*{Silhouette-based Differentiable Rendering and Pose Estimation}
As 2D DSA imaging generates two images from two X-Ray projectors, one for each image $\mathbf{I}_\text{AP}$ and $\mathbf{I}_{L}$, we can formulate our optimization problem as simultaneously minimizing the difference between the two DSA images and their corresponding rendered images: 
\begin{equation}
\widehat{\mathbf{P}_\text{AP}}; \widehat{\mathbf{P}_\text{L}} = \underset{\mathbf{P}_\text{AP}, \mathbf{P}_\text{L}}{\mathrm{argmin}} \; \frac{1}{N} \sum_{i=1}^{N} \bigg (\Big \Vert \mathbf{I}_\text{AP}  - f\big(\mathbf{P}_\text{AP}|\mathbf{M} \big)\Big \Vert_2^2 + \Big \Vert \mathbf{I}_\text{L}  - f\big(\mathbf{P}_\text{L}|\mathbf{M}\big)\Big \Vert_2^2 \bigg )
\label{eq:loss1}
\end{equation}
We want to solve Eq. \ref{eq:loss1} using only one X-Ray projector, for instance, $\mathbf{P}_\text{L}$. 
We model the X-Ray beam projector as $\mathbf{P} = \mathbf{K}[\mathbf{R}|\mathbf{t}]$, where $\mathbf{K}$ is the matrix of intrinsic parameters, $\mathbf{R}  \in SO(3)$ is the rotation matrix and $\mathbf{t} \in \mathbb{R}^3$ is the translation vector. 
Given an axis of rotation $\mathbf{a} = \begin{pmatrix} a_x & a_y & a_z \end{pmatrix}^\top$ extracted from $\mathbf{R}_\text{L}$, so that $\mathbf{a} = \mathbf{R}_\text{L}[:,1]$ and a rotation angle $\theta$, we can calculate a rotation matrix $\mathbf{W}$ using Rodrigues' Rotation Formula as:
\begin{equation}
\mathbf{W} = \mathbb{I} + \sin\left(\theta\right) [\mathbf{a}]_\times + \left(1 - \cos\left(\theta\right)\right) [\mathbf{a}]_\times^2
\end{equation}
Assuming the X-Ray projectors are perpendicular, i.e.: $\theta = 90^{\circ} = \pi/2$ radians, this simplifies to:
\begin{equation}
\mathbf{W} = \mathbb{I} + [\mathbf{a}]_\times + [\mathbf{a}]_\times^2
\end{equation}
where $\mathbb{I}$ is the identity matrix and $[\mathbf{a}]_\times$ is a skew-symmetric matrix. 
Therefore, the rotation matrix $\mathbf{W}$ for a 90-degree rotation around the axis $\mathbf{a}$ is:
\begin{equation}
\mathbf{W} = \begin{pmatrix}
1 - a_y^2 - a_z^2 & a_x a_y - a_z & a_x a_z + a_y \\
a_x a_y + a_z & 1 - a_x^2 - a_z^2 & a_y a_z - a_x \\
a_x a_z - a_y & a_y a_z + a_x & 1 - a_x^2 - a_y^2
\end{pmatrix}
\end{equation}

Assuming the projector has similar intrinsic parameters $\mathbf{K}$, we can use $\mathbf{W}$ we can re-write Eq. \ref{eq:loss1} using solely the lateral rotation and translation $[\mathbf{R}_\text{L}|\mathbf{t}]_\text{L}$ as follows: 
\begin{equation}
\underset{[\mathbf{R}_\text{L}|\mathbf{t}_\text{L}]}{\mathrm{argmin}} \; \frac{1}{N} \sum_{i=1}^{N} \bigg (\lambda_1 \Big \Vert \mathbf{I}_\text{L}  - f\big(\mathbf{K}[\mathbf{R}_\text{L}|\mathbf{t}_\text{L}]|\mathbf{M} \big)\Big \Vert_2^2 + \lambda_2 \Big \Vert \mathbf{I}_\text{L}  - f\big(\mathbf{K}[\mathbf{W}\mathbf{R}_\text{L}|\mathbf{t}_\text{L}]|\mathbf{M}\big)\Big \Vert_2^2 \bigg )
\label{eq:loss2}
\end{equation}

The differentiable nature of $f$ allows computing $\frac{\partial \mathcal{L}}{\partial \mathbf{P}_{L}}$, thereby enabling an iterative pose refinement via gradient descent. 
The parameters $\lambda_1$ and $\lambda_2$ control the weights of each X-Ray projector so that the anteroposterior images are incorporated progressively into the iterative pose estimation.  
In practice, we rely on segmentations of $\mathbf{I}_\text{L}$ that perform well with vessels when compared with the rasterized mesh $\mathbf{M}$.

\section{Results}
We first tested our method by a registration of synthetic images. We derived a 3D mesh of the brain artery from a 3D DSA image from this dataset \cite{mitrovic_spiclin_3D-2D-GS-CA} using Otsu's method to segment the 3D DSA image, and manually estimated a ground-truth X-Ray position. We randomly initialized our system's X-ray position(s) to generate 2D silhouette(s) based on their position(s) and optimized their pose estimation iteratively using gradient descent.

We measured the impact of using both images in estimating the pose. We tested several configurations: AP then LAT, AP alone, AP+LAT simultaneously, LAT then AP, and LAT alone, and tested their efficacy based on the 6 Degrees-of-Freedom (6D) pose error and the Average Distance of Model Points (ADD) error. The 6D error measures the offset rotation and translation between the estimated and ground-truth X-ray positions, while the ADD error measures the average Euclidean distance between the corresponding vertices of each of the meshes of the estimated and ground-truth pose.

\begin{table}[t]
\caption{6D and ADD Pose Estimation Errors}\label{tab:6d_error}
\centering
\begin{tabular}{lccc}
\toprule
Configurations   & Rotation Error [$^\circ$] & Translation Error [mm] & ADD Error [mm] \\
\midrule
\rowcolor{gray!10}
AP, then LAT    & 0.0248            & 0.0670     & 0.0105          \\
AP              & 18.8452           & 53.8050    & 8.9186          \\
\rowcolor{gray!10}
AP+LAT simultaneous & 0.0319           & 0.0657     & 0.0188          \\
LAT, then AP    & 0.0132            & 0.0294     & 0.0195          \\
\rowcolor{gray!10}
LAT             & 0.0133            & 0.0288     & 0.0200          \\
\botrule
\end{tabular}
\label{tab:6d_error}
\end{table}

\begin{figure}[h]
    \centering
    \href{https://github.com/taewoonglee17/TwoViewsDSAReg/blob/main/figs/3D_Demo_DSA_Registration_synthetic.gif}{
        \includegraphics[clip, trim=0 0 0 280, width=0.35\textwidth]{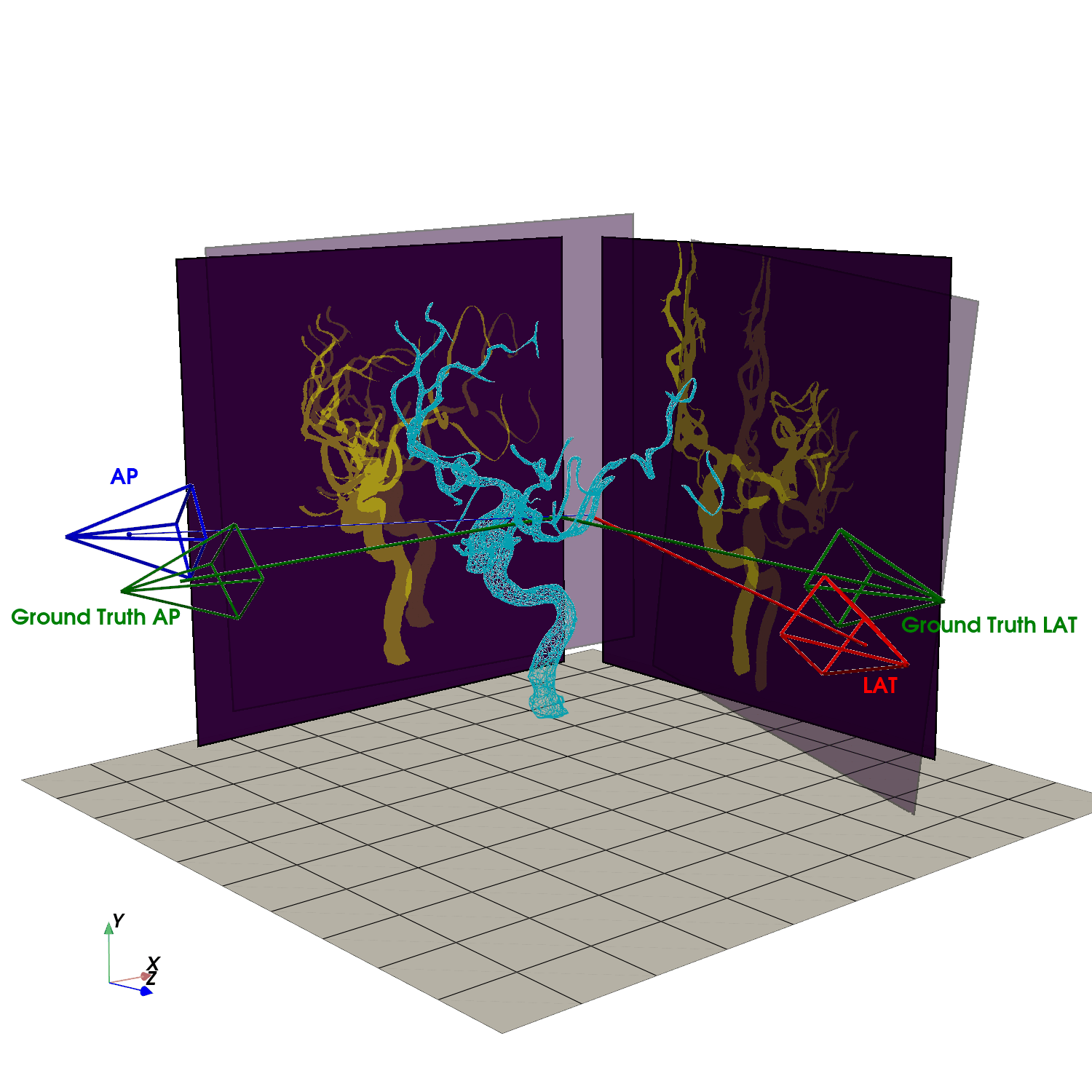}%
        \hfill
        \includegraphics[width=0.3\textwidth]{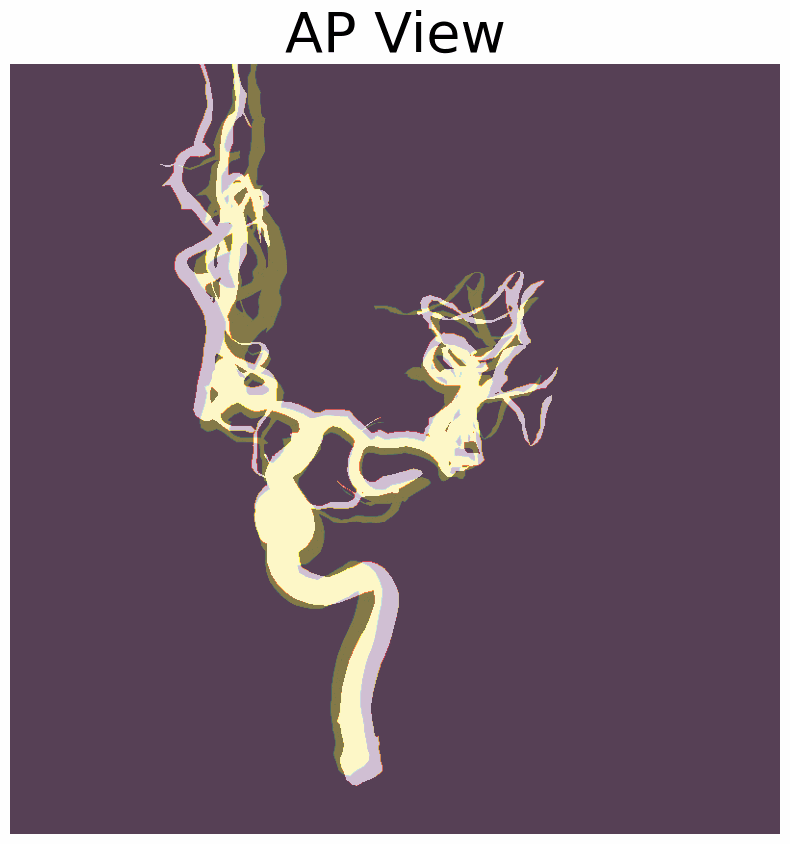}%
        \hfill
        \includegraphics[width=0.3\textwidth]{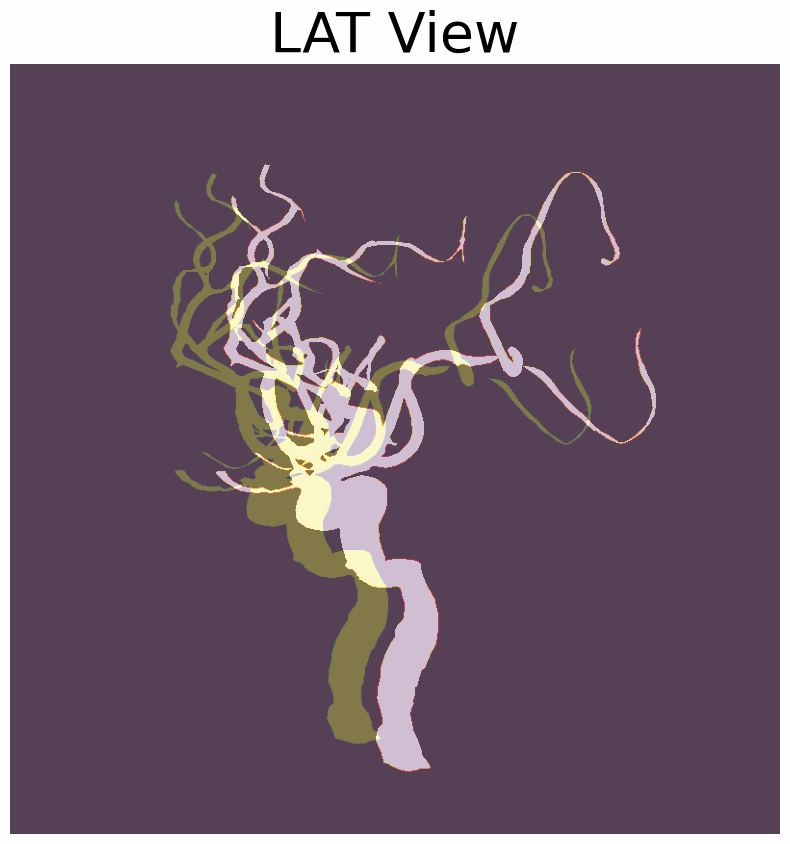}\\ 
        \includegraphics[clip, trim=0 0 0 280, width=0.35\textwidth]{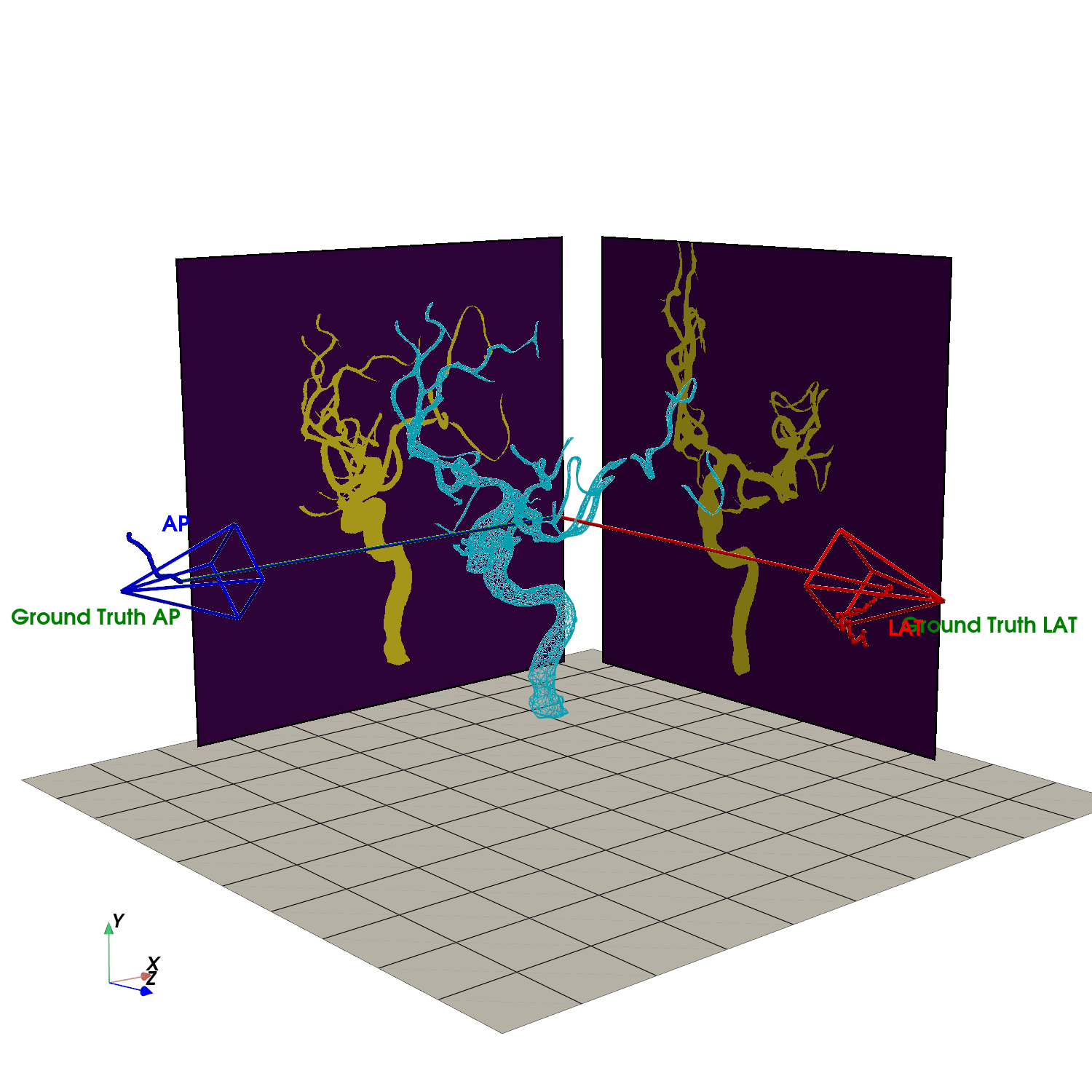}%
        \hfill
        \includegraphics[width=0.3\textwidth]{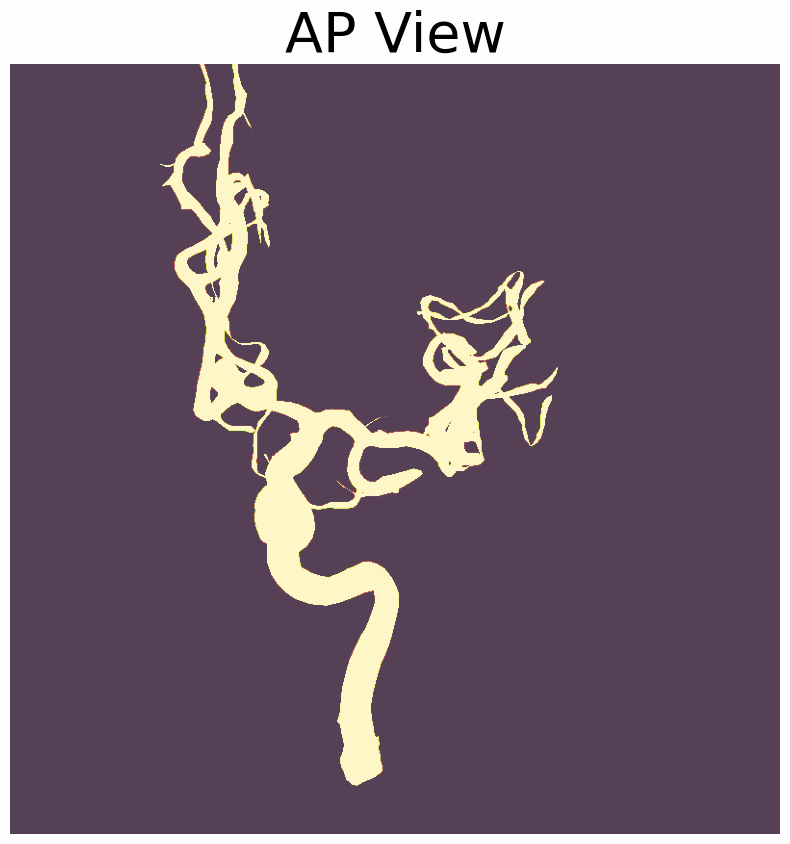}%
        \hfill
        \includegraphics[width=0.3\textwidth]{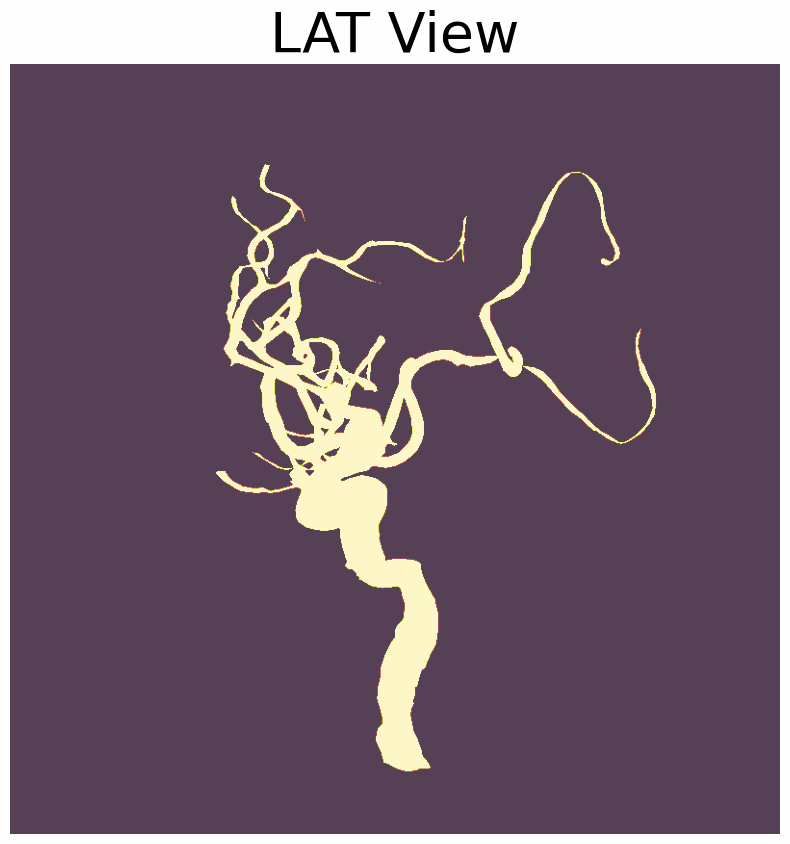}%
    }
\caption{First (top) and final (bottom) iterations of the registration process using the two AP and LAT images for synthetic data. \textbf{Click the image to play the video in a browser.}}
\label{fig:restults}
\end{figure}

Our results in Table \ref{tab:6d_error} indicate that utilizing two viewing angles consistently outperforms relying on a single view. While the LAT view alone demonstrates performance comparable to the combined AP+LAT views, it shows a notable improvement over the AP view alone, and relying on a single view inherently introduces the potential for greater errors. In contrast, the AP+LAT configuration consistently achieves stable, low error levels — particularly for ADD error — regardless of the acquisition sequence (AP first, LAT first, or simultaneously).
Figure \ref{fig:restults} shows an example of our method applied to synthetic data.

Furthermore, we tested our approach with real DSA images using the same dataset \cite{mitrovic_spiclin_3D-2D-GS-CA}. A qualitative assessment of the registration and pose estimation is provided in Figure \ref{fig:overview}.

\section{Conclusion}
We proposed a silhouette-based differentiable rendering approach for 3D/2D angiographic registration using two viewpoints, demonstrating robust pose estimation and accurate reconstruction on both real and synthetic datasets. Future work includes integrating state-of-the-art segmentation techniques for real-world data and optimizing performance through parallel initialization. Additionally, we aim to parameterize the angle between the AP and LAT projectors rather than a fixing heuristic modification that will better replicate real-world imperfections. Finally, we plan to implement a more flexible and adaptive rasterizer capable of dynamically adjusting rasterization levels based on blood vessel thickness. This enhancement seeks to address potential optimization imbalances, where thinner vessels may otherwise be underweighted relative to thicker ones.

\section*{Acknowledgements}
This project was funded in part by the National Institutes of Health grants: R01EB034223, R03EB033910, and K25EB035166.

\bibliography{sn-bibliography}
\end{document}